\def\year{2021}\relax
\documentclass[letterpaper]{article} % DO NOT CHANGE THIS
\usepackage{aaai21}  % DO NOT CHANGE THIS
\usepackage{times}  % DO NOT CHANGE THIS
\usepackage{helvet} % DO NOT CHANGE THIS
\usepackage{courier}  % DO NOT CHANGE THIS
\usepackage[hyphens]{url}  % DO NOT CHANGE THIS
\usepackage{graphicx} % DO NOT CHANGE THIS
\usepackage{graphicx}  % DO NOT CHANGE THIS
\usepackage{booktabs}
\usepackage{multirow}
\usepackage{kotex}
\usepackage[usenames,dvipsnames]{xcolor}
\usepackage{soul}
\usepackage{algorithm}
\usepackage{algpseudocode}
\newcommand{\etal}{\textit{et al}.}
\usepackage{enumitem}
\usepackage{amsmath}

\title{AutoLR: Layer-wise Pruning and Auto-tuning of Learning Rates \\in Fine-tuning of Deep Networks} 
\author{
  Youngmin Ro\textsuperscript{1,2}
  \,
  Jin Young Choi\textsuperscript{1}
  \\
  \small{\texttt{youngmin.ro@samsung.com},\; 
 \texttt{jychoi@snu.ac.kr}} \\
  \textsuperscript{1}Department of ECE, ASRI, Seoul National University, Korea \\
\textsuperscript{2}Samsung SDS, Korea\\
}% \textsuperscript{\rm 1}\thanks{Primarily Mike Hamilton of the Live Oak Press, LLC, with help from the AAAI Publications Committee}\\ \Large \textbf{AAAI Style Contributions by
\begin{document}

\maketitle

\begin{abstract}
Existing fine-tuning methods use a single learning rate over all layers. In this paper, first, we discuss that trends of layer-wise weight variations by fine-tuning  using a single learning rate do not match the well-known notion that lower-level layers extract general features and higher-level layers extract specific features.
Based on our discussion, we propose an algorithm that improves fine-tuning performance and reduces network complexity through layer-wise pruning and auto-tuning of layer-wise learning rates.
The proposed algorithm has verified the effectiveness by achieving state-of-the-art performance on the image retrieval benchmark datasets (CUB-200, Cars-196, Stanford online product, and Inshop). Code is available at \url{https://github.com/youngminPIL/AutoLR}.

\end{abstract}

%%%%%%%%% BODY TEXT
%\input{1-Introduction}
\section{Introduction}
The ability to collect large amounts of data has allowed deep learning to advance dramatically over the last decade. In particular, deep neural network architectures have evolved by targeting large datasets such as ImageNet~\cite{ref:imagenet}.
However, in actual computer vision applications, large amounts of data such as those contained ImageNet, cannot be easily obtained.
Thus, for applications such as image retrieval or fine-grained classification, for which only small datasets are available, the use of pre-trained networks has become essential~\cite{ref:RA_CNN,ref:MA_CNN,ref:Stochastic_Class_Based,ref:spot-tune,ref:Multi-Similarity}. 
To utilize the pre-trained network for a target task, we use a fine-tuning scheme that initially takes pre-trained weights and adjusts them in whole or in part.

The performance of the fine-tuning depends highly on the following factors: similarity between the source and target tasks~\cite{ref:Factor_transferability_distance,ref:Large_FGVC}, choice of deep network model~\cite{ref:do_better_imagenet}, and tuning strategy~\cite{ref:full_or_fine-tune_medical,ref:spot-tune,ref:ro_rollback}. From among these factors, our paper focuses on enhancement of tuning strategy. Many efforts have been made to enhance the capability of tuning the network for a given target task.
Partial tuning~\cite{ref:full_or_fine-tune_medical} has been proposed to tune only the higher-level layers. Weight-reverting methods were proposed in some studies~\cite{ref:spot-tune,ref:ro_rollback}, where the tuned weights are reverted to the initial (pre-trained) weights during fine-tuning. In addition, some researchers adjusted the learning rate (LR) periodically to ensure efficient learning by adjusting it to follow a triangular 
waveform~\cite{ref:cyclical}, and even proposing an exponentially decaying shape of LR~\cite{ref:warmSGD}. However, most existing fine-tuning methods adopt a single LR regardless of layers.   

%%%%%%%%%%%%%%%%%%%%%%%%%
\begin{figure}[t]
\centering
{\includegraphics[width=9cm]{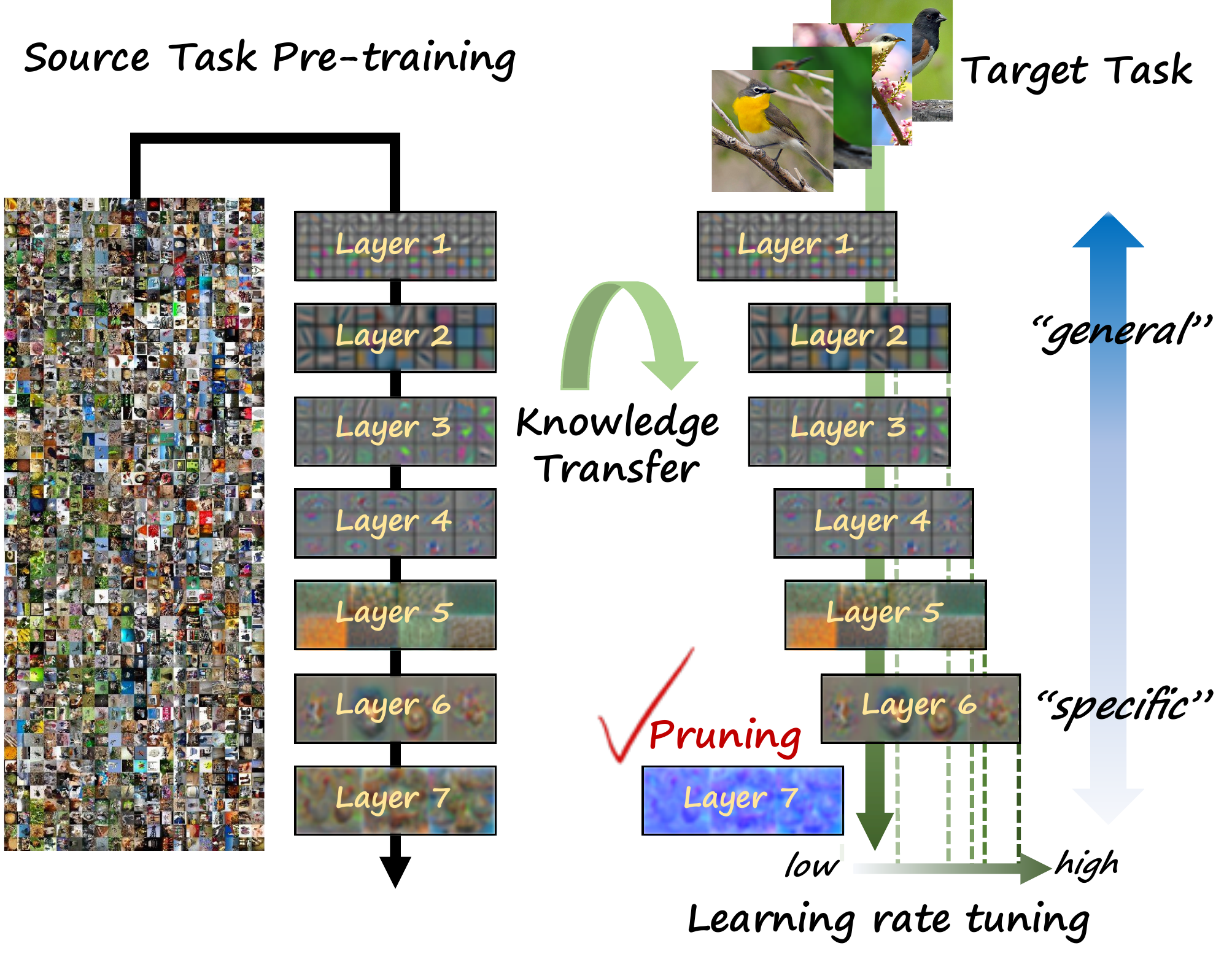}}
 \vspace{-0.5cm}
    \caption{The conceptual figure of the proposed algorithm: Conducting layer-wise pruning and auto-tuning of layer-wise learning rates on the target task according to role of each layer.}
    \label{fig:cover}
    \vspace{-0.4cm}
\end{figure}

%%%%%%%%%%%%%%%%%%%%%%%%%%

In this paper, instead of using a single LR, we propose an algorithm for layer-wise auto-tuning of LRs where the LR in each layer is automatically tuned according to its role. In addition, we propose a layer-wise pruning algorithm that removes layers according to the usefulness of each pre-trained layer to the new task as illustrated in Figure~\ref{fig:cover}. 
Our work is inspired by our observation that the trends of layer-wise weight variations by conventional fine-tuning using a single LR, contradict previous studies~\cite{ref:how_transferable,ref:visualizing_conv}, which claim that low-level layers extract general features while high-level layers extract specific features. 
Based on our observation, we establish two hypotheses and validate them empirically via preliminary experiments. Following the validation of the hypotheses, we develop algorithms for layer-wise pruning and auto-tuning of layer-wise LRs.

Through ablation experiments on four image retrieval benchmark datasets, it is verified that the proposed method consistently improve the retrieval performance.
By providing visualization results, we help to understand the mechanism of the proposed method.
In addition, the superiority of our method as a fine-tuning strategy is demonstrated through comparative experiments with the existing LR setting algorithms.
Lastly, with regard to four benchmark datasets of image retrieval, our method outperforms the existing state-of-the-art methods.

\section{Related Work}
Fine-tuning is a kind of transfer learning and is used for tuning of pre-trained parametric model. 
In fine-tuning, the similarity between the source task and the target task is also important as in transfer learning. Regarding the studies on the similarity, 
Azizpour~\etal~\shortcite{ref:Factor_transferability_distance} has suggested factors to transfer knowledge well considering the similarity between target and source tasks. And
Cui~\etal~\shortcite{ref:Large_FGVC} proposed a method to promote knowledge transfer using similarity between multiple tasks. 
In the computer vision fields, 
ImageNet~\cite{ref:imagenet} dataset has been widely used for a source task. Hence lots of target tasks have been handled by using deep network models pre-trained by using ImageNet. 
In Kornblith~\etal~\shortcite{ref:do_better_imagenet}, it has been claimed that there is a positive correlation between ImageNet and most target tasks regardless of deep networks. This implies that ImageNet includes huge amounts of image data covering most image-related tasks. Hence in our paper, we will adopt ResNet-50 pre-trained using ImageNet to validate our algorithm. 

In our paper, we are motivated from the role of each layer in fine-tuning a deep network. Regarding the studies on the roles of hidden layers,
Yosinski~\etal~\shortcite{ref:how_transferable} conducted empirical study to quantify the degree of generality/specificity of each layer in deep networks. And  
Zeiler~\etal~\shortcite{ref:visualizing_conv} visualized features in hidden layers to analyze generality/specificity in the layers. 
Through the studies~\cite{ref:how_transferable,ref:visualizing_conv}, they claim that the low-level layers extract general features and the high-level layers extract specific features in a deep network. This claim provides insight into our algorithm.

There have been similar works to ours that exploits the role of each layer.
In~\cite{ref:full_or_fine-tune_medical}, Tajbakhsh~\etal have shown that tuning only a few high-level layers is more effective than tuning all layers.
Guo~\etal~\shortcite{ref:spot-tune} proposed an auxiliary policy network that decides whether to use the pre-trained weights or fine-tune them in layer-wise manner for each instance.
Ro~\etal~\shortcite{ref:ro_rollback} proposed a rollback scheme that returns a part of weights to their pre-trained state to improve performance. However, these works adopt the fixed learning rate settings unlike our method.
In addition, the approach of `learning to learn'~\cite{li2017meta,pmlr-v80-ren18a} aims to find the optimal update of network weights. This approach directly finds the optimal initialization and update of the network weights unlike our approach to efficiently update the pre-trained weights via auto-tuning layer-wise learning rate.

Similarly to ours, there are studies that adjust the learning rate periodically during learning process. 
Smith~\shortcite{ref:cyclical} suggested a way to adjust the learning rate in a triangular form that linearly reduces learning rate and then grows it again.
Similarly Loshchilov~\etal~\shortcite{ref:warmSGD} has also proposed an exponentially decaying and restarting the learning rate for a certain period of time.
However, the existing methods for on-line tuning of the learning rate use single learning rate over all the layers. In our work, we aims to develop an algorithm to auto-tune the layer-wise learning rates depending on the role of each layer.

%%%%%%%%%%%%%%%%%%%%
\begin{figure*}[t!]
\centering
{\includegraphics[width=18cm]{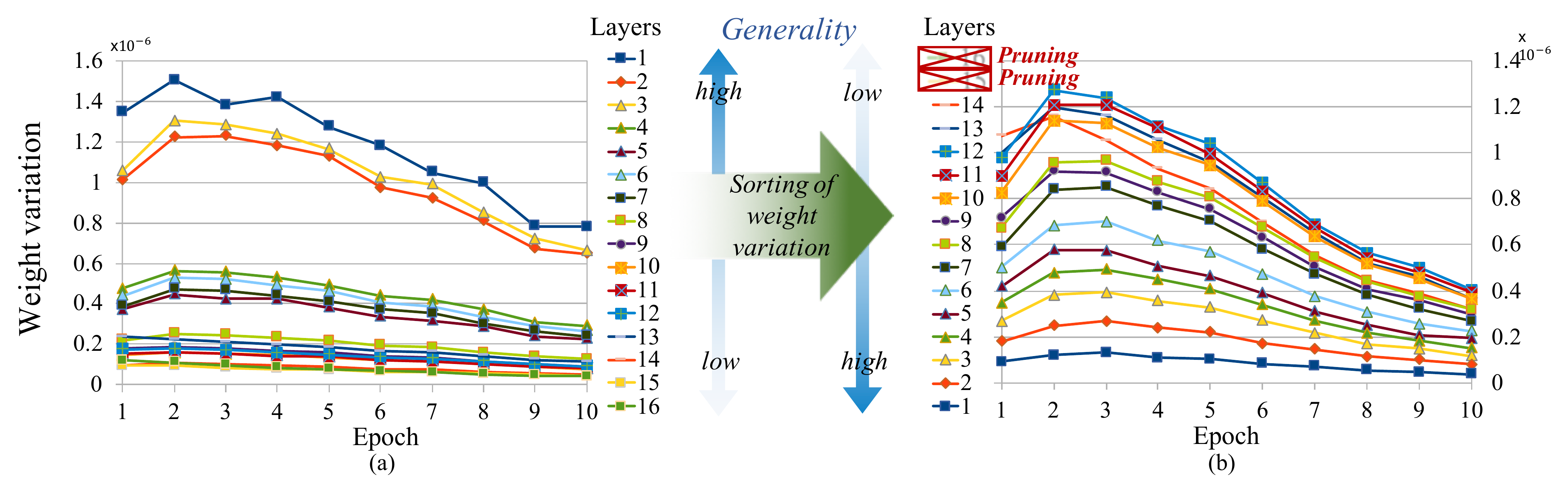}}
    \hfill
    \vspace{-1.0cm}
    \caption{The trend of the weight variation between two adjacent epochs. In the result of the conventional fine-tuning of single LR (a), it is observed that the variations of the higher-level layers are small and the lower-level layers are large. (b) The trend after pruning two highest-level layers and tuning layer-wise LRs for sorting of weight variation. 
    % by trial-and-error.
    After pruning and layer-wise LR tuning, the performance improved significantly (In Table~\ref{table:ablation_pre}).
    }
    \label{fig:preliminary_CUB}
\end{figure*}
%\input{input_fig/preliminary_CUB.tex}
%%%%%%%%%%%%%%%%%%%%

\section{Proposed Method}
The purpose of this paper is to devise an algorithm that automatically sets the appropriate learning rate (LR) for each layer after removing layers that do not contribute to the learning results in the fine-tuning process.
We first discuss our findings observed in the conventional fine-tuning process and then analyze the effects of the layer-wise pruning or layer-wise tuning of LRs through preliminary experiments. Then, we derive an approximate relationship between the weight variation of each layer and the LR. Finally, we propose an algorithm for the layer-wise pruning and auto-tuning of LR in each layer.

%%%%% 11월 6일 transferrablity 백업함.

\subsection{Layer-wise Weight Variations in Fine-tuning}
\label{effect_learning_rate}
This paper is inspired by previous studies~\cite{ref:how_transferable,ref:visualizing_conv,ref:Factor_transferability_distance}.
These studies have demonstrated empirically and qualitatively that lower-level layers extract general features and higher-level layers extract specific features.
Based on these results, we build two hypotheses: \\
{\bf Hypothesis 1:} {\it The pre-trained high-level layers may not be helpful to a new target task because they are specific to the source task.} \\
{\bf Hypothesis 2:} {\it The weight variations of pre-trained low-level layers would be small because they are generally valid for most tasks, whereas those in high-level layers would be large because they should adopt themselves to a new task specifically.}   

To clarify these hypotheses and support our idea, we define the weight variation of $k$-th layers as
\begin{eqnarray}
\label{eq:weight_variation}
v_{t}^{k} = \frac{1}{n_k}\| \Delta w^k_{t}\|, 
\end{eqnarray}
where $\Delta w^k_{t} = w^k_{t} - w^k_{t-1}$ denotes the amount of weight changes in the $k$-th layer during $t$-th epoch, and $n_k$ is the number of weights in the $k$-th layer.

First, we investigate the trends of layer-wise weight variations in the conventional fine-tuning process.  
The investigation has been conducted on a retrieval dataset CUB-200-2011~\cite{ref:CUB-2011} with ResNet-50~\cite{ref:resnet}.
Figure~\ref{fig:preliminary_CUB} shows the trends of weight variation between two consecutive epochs.
`Layer' denotes a residual block in Resnet-50, where `Layer 1' indicates the lowest layer to the input layer and thus `Layer 16' becomes the highest layer. From the results shown in
Figure~\ref{fig:preliminary_CUB}-(a), which shows the trends of layer-wise weight variations by the conventional fine-tuning with single LR, we have observed two interesting points described in the following two paragraphs. These observations support the key ideas of our paper.

The first point is that the weight variations in high-level layers are too small from fine-tuning, as shown in Figure~\ref{fig:preliminary_CUB}-(a).  
Based on Hypothesis 1, the high-level layer with small weight variation may not contribute to the new task learning. To support this claim, we have conducted a layer-wise pruning experiments in which the high-level layers were removed one by one from the highest layer. The performance variations by the pruning are shown in  
Table~\ref{table:ablation_pre}.  
Interestingly, the pruning Layer 15, 16 improves the performance, which implies that Layers 15 and 16 might not be helpful to the target task.
The pruning of layers until Layer 14 does degrades the performance, which means the layers below Layer 14 are helpful to the new task.
This preliminary result supports Hypothesis 1 and can be motivation for our simple but efficient layer-wise pruning scheme.

The second point is that the trends in weight variations by traditional fine-tuning do not match Hypothesis 2, as shown in Figure~\ref{fig:preliminary_CUB}-(a). 
This implies the traditional fine-tuning destroys the general features of the pre-trained network by large variations in low-level layers and cannot promote the high-level layer to adapt itself to the new task. 
We believe this result is due to the LR assigned by a single value regardless of layers. To verify this, we have conducted a preliminary experiment by adjusting layer-wise LRs empirically. The layer-wise adjustment of LRs gives a significant improvement, 
as shown in rows 5, 6, and 7 of Table~\ref{table:ablation_pre}. 
Figure~\ref{fig:preliminary_CUB}-(b) shows the order of layer-wise weight variations for row 7 (Pruning 15, 16$^\star$). Interestingly, we can see the order for row 7 meets Hypothesis 2. This result validates Hypothesis 2 that can be utilized to design an auto-tuning scheme for layer-wise LRs.

%%%%%%%%%%%%%%%%%%%%%%%%%%%%%%%%%%%%%%%%%%%%%%%%%%%%%%%%%%%%%%%%%%%%%%%%%%%
\begin{table}[t]
% \resizebox{0.5\textwidth}{!}{
\caption{The Recall@\textit{K} score results of preliminary experiment for layer-wise pruning and layer-wise adjusting of LRs. In $\dagger$, the LRs in Layers 1,2,3 were reduced to 1/10.
In $\ddagger$, the LRs in Layer 13,14 were increased by 10 times.
In $\star$, the LR in all layers were empirically adjusted to meet Hypothesis 2.} %by trial and error way.}
\label{table:ablation_pre}
\vspace{0.2cm}
\centering
\begin{tabular}{lcccc}
\toprule
  & R@1  		& R@2		& R@4	& R@8  \\ \midrule
Original        & 63.49		& 75.03         & 84.00         & 90.58   \\ 
Pruning 16      & 66.95		& 77.63 		& 86.39 		&92.00\\ 
Pruning 15, 16 		 & \underline{67.00} 	& \underline{78.49	}	& \underline{86.85}		&\underline{92.07} \\ 
Pruning 14, 15, 16	 & 51.00	    & 64.16 		& 75.44 		&85.94 \\\midrule
Pruning 15, 16$^\dagger$ 		 & 67.15 		& 78.61			& 86.83			&92.08 \\
Pruning 15, 16$^\ddagger$		 & 67.29 		& 79.15		& 87.36		&92.66 \\ \midrule
Pruning 15, 16$^\star$	 & \textbf{68.33}		& \textbf{79.88} 		& \textbf{87.69} 	&\textbf{92.71} \\

\bottomrule
\end{tabular}	%}
\end{table}

%top1:0.927257 top5:0.977435 top10:0.986639 mAP:0.818032

% remain layers  & top-1    & mAP      \\ \midrule
% Block1 + FC & 0.90.1425 & 0.75.1156 \\
% Block2 + FC & 0.90.2613 & 0.75.9104 \\
% Block3 + FC & 0.90.2613 & 0.75.6990 \\
% Block4 + FC & 0.90.1425 & 0.74.8464 \\
% Block5 + FC & 0.89.2815 & 0.73.3921 \\ \bottomrule
%%%%%%%%%%%%%%%%%%%%%%%%%%%%%%%%%%%%%%%%%%%%%%%%%%%%%%%%%%%%%%%%%%%%%%%%%%%%%%%%%%%%%%%

Based on Hypotheses 1 and 2, we aim to develop an algorithm for the layer-wise pruning and auto tuning of LR (AutoLR). In particular, the AutoLR is designed to achieve the order of layer-wise weight variations in every epoch that meets Hypothesis 2, that is, 
\begin{align}
\label{eq:ordering_of_variation}
v^{1}_{t}\le v^{2}_{t} \le
\cdots \le v^{K}_{t}.    
\end{align}

\subsection{Weight Variation and LR}
\label{weight_learning_rate}
Our key idea is to control the weight variation of each layer by tuning the LR of each layer in every epoch. In this section, we derive a relationship between the LR and the weight variation, which will be used in the auto-tuning scheme for LRs according to the layers.

The weights in the $k$-th layer are updated during $l$-th iteration for a randomly chosen mini-batch by the stochastic gradient descent rule with the momentum as follows:
\begin{align}
\label{eq:SGD_update_rule}
\Delta w^k_{t,l} &= \rho \Delta w^k_{t,l-1} -\eta^{k} \nabla{\mathcal{L}(w^k_{t,l-1})}, %+\beta_{t,l},
\end{align}
where $\Delta w^k_{t,l}=w^k_{t,l}-w^k_{t,l-1}$,  ${\mathcal{L}(w^k_{t,l-1})}$ denotes the loss for $w^k_{t,l-1}$, $\rho$ is a momentum coefficient, $l$ is the iteration index, $t$ is the epoch index, and $\eta^{k}$ is LR of $k$-th layer. Note that $\eta^{k}$ is adjusted every epoch in our AutoLR scheme, hence $\eta^{k}$ is constant for all iterations during an epoch.
Therefore, by applying \eqref{eq:SGD_update_rule} recursively, we can obtain  
\begin{align}
\Delta{w^k_{t,l}} &= -\eta^{k}[\nabla{\mathcal{L}(w^k_{t,l-1})} + \rho \nabla{\mathcal{L}(w^k_{t,l-2})} \notag\\ 
+&\rho^2 \nabla{\mathcal{L}(w^k_{t,l-3})} +\rho^3 \nabla{\mathcal{L}(w^k_{t,l-4})} + \cdots ]. 
\end{align}
Define
\begin{align}
\label{Lacc}
\nabla{\mathcal{L}_{acc} (w^k_{t})} &
=\sum_{l=1}^{L} [\nabla{\mathcal{L}(w^k_{t,l-1})} + \rho \nabla{\mathcal{L}(w^k_{t,l-2})}\notag \\ 
+&\rho^2 \nabla{\mathcal{L}(w^k_{t,l-3})} +\rho^3 \nabla{\mathcal{L}(w^k_{t,l-4})} + \cdots ],
\end{align}
where $L$ is the number of mini-batches and $\nabla{\mathcal{L}(w^k_{t,l})}=0$ for $l\le0$. 
Then 
% \ro{epoch-wise weight changes of k-th layer}
\begin{align}
\Delta{w^k_{t}}=\sum_{l=1}^L\Delta{w^k_{t,l}}=-\eta^{k}\nabla{\mathcal{L}_{acc}(w^k_{t})}, 
\end{align}
which leads to 
\begin{align}
\label{eq:norm_weight_variation}
\| \Delta{w^k_{t}}\| = {\eta^{k}}\|\nabla{\mathcal{L}_{acc}(w^k_{t})}\|.
\end{align}
From \eqref{eq:weight_variation} and \eqref{eq:norm_weight_variation}, 
\begin{eqnarray}
\label{eq:normalized_weight_variation}
v_{t}^{k} = \frac{\eta^{k}}{n_k} \|\nabla{\mathcal{L}_{acc}(w^k_{t})}\|. 
\end{eqnarray}
In the next section, this equation is used for the AutoLR scheme.
Note that $\|\nabla{\mathcal{L}_{acc}(w^k_{t})}\|$ does not explicitly depend on $\eta_k$ as shown in  (\ref{Lacc}). Actually, the variation $\|\nabla{\mathcal{L}_{acc}(w^k_{t})}\|$ has been observed to be negligible (see Appendix B of the supplementary material\footnote{\url{https://github.com/youngminPIL/AutoLR}}).
%\ro{however, it is observed that they do not depend on $\eta_k$ as much.}
% which arises from the vanishing of the high order of $\eta_k$ in the terms.
In AutoLR scheme, hence, we ignore the variation of $\|\nabla{\mathcal{L}_{acc}(w^k_{t})}\|$ during iterations in each epoch.
%\ro{as $\eta_k$ changes during an epoch.} 
We apply the equation (\ref{eq:normalized_weight_variation}) to our AutoLR scheme repeatedly until we get the goal in \eqref{eq:ordering_of_variation}.

\subsection{Layer-wise Pruning and Auto-tuning of LR}
\label{tuning_learning_rate}
Based on the results analyzed above, we propose an algorithm to prune layers that are not helpful to the target task. Then we also propose an algorithm (AutoLR) to  automatically adjust the LR of each layer so that the order of the weight variation size of each layer is consistent with \eqref{eq:ordering_of_variation}.
AutoLR is divided into two parts: setting the target weight variation and tuning by adjusting the LR accordingly.

%%%%%%%%%%%
\newcommand\tab[1][1.2cm]{\hspace*{#1}}
\renewcommand{\algorithmicrequire}{\textbf{Pruning:}}
\renewcommand{\algorithmicensure}{\textbf{Notation:}}
\renewcommand{\algorithmicprocedure}{\textbf{Pruning:}}
\algtext*{EndWhile}
\algtext*{EndIf}

\algdef{SE}[DOWHILE]{Do}{doWhile}{\algorithmicdo}[1]{\algorithmicwhile\ #1}

\begin{algorithm} [h]
\caption{Layer-wise Pruning}\label{alg:pruning}
\small{
\begin{algorithmic}[1]
\Ensure 
\Statex $\eta$ : a single learning rate for all layers
\Statex \textit{network} : network with weight parameters
\Statex \textit{score} : performance of  current network
\Statex \textit{best score} : best performance of previous networks
\Require 
\State  \textit{network} $\gets$ \textit{pre-trained network}
\State Set $\eta$ to the all layers of network 
\State Set {\it best score}  $=0$
\State Fine-tune \textit{network} 
\State {\it score} $\gets$ performance of \textit{network}
   \While {\textit{score} $\geq$ \textit{best score}}
    \State {\it best score} $\gets$ {\it score}
        \State  \textit{network} $\gets$ \textit{pre-trained network}
        \State Prune the highest-level layer of {\it network}
        \State Fine-tune \textit{network} %for target dataset
        \State {\it score} $\gets$ performance of \textit{network}
        \EndWhile 
   \end{algorithmic}}
\end{algorithm}

%%%%%%%%%%%%%%

\subsubsection{Layer-wise pruning rule} 
\label{Pruning Rule}
Before applying layer-wise AutoLR, low-contributed high-level layers are pruned by the procedure in Algorithm 
\ref{alg:pruning}. In the pruning procedure, we fine-tune a network on a target task using the traditional fine-tuning scheme with a layer-wise fixed LR.

\subsubsection{Setting target weight variations}
We need to renew the target weight variations that satisfy the goal in \eqref{eq:ordering_of_variation} in each epoch of the learning process.
We design two formulas to set the renewed target weight variation depending on {\it sorting quality} that denotes how well the current weight variations satisfy the goal in \eqref{eq:ordering_of_variation}.
To this end, we measure the {\it sorting quality} defined as follows:
\begin{align}
sorting~ quality=1-\frac{2}{K^2}\sum_{k=1}^K|k-\sigma(v_t^{k})|,
\end{align}
where $\sigma(v_t^{k})$ is a function that maps $v_t^{k}$ to its ranking in ascending order, and $\frac{2}{K^2}$ is the scale factor that enforces the sorting quality to be scaled within 0 and 1.

The initial target weight variation is set as follows:
\begin{align}
d_t = \frac{1}{K-1}\left({\beta}\max_{{1 \le k \le K}}v_{t}^{(k)} - {\alpha}\min_{{1 \le k \le K}}v_{t}^{(k)}\right)\\
\bar{v}_{t}^{(k)} \gets \min_{1 \le i \le K}v_{t}^{(k)} + (k-1)d_t, \text{ }k = 1, \cdots, K.
\end{align}
where the $\alpha$ and $\beta$ are the hyper-parameter to set a range of target weight variation.

If the {\it sorting quality} is below a pre-defined threshold $\tau_{s}$, the target weight variation requires partial modifications. 
To prevent the renewed target sorting from shifting to one side, the modification starts from the center ($\check{k}$) and moves to both ends as follows:
\begin{align}
\text{For }k =& \check{k}, \check{k}+1, \cdots, K\notag\\ 
\bar{v}_{t}^{(k)}& \gets 
\begin{cases}
v_{t}^{(k)} &              k = \check{k} \text{  or  }\bar{v}_{t}^{(k-1)} \le v_{t}^{(k)}\\
\bar{v}_{t}^{(k-1)} + d_t & \text{otherwise},
\end{cases} \\
\text{For }k =& \check{k}-1, \check{k}-2, \cdots, 1 \notag\\
\bar{v}_{t}^{(k)}& \gets 
\begin{cases}
v_{t}^{(k)} &             \bar{v}_{t}^{(k+1)} \ge v_{t}^{(k)}\\
\bar{v}_{t}^{(k+1)} - d_t & \text{otherwise}.
\end{cases} 
\end{align}

\subsubsection{Renewing LR}  

To get a new LR $\bar{\eta}_{t}^{k}$ which produces the renewed target weight variance $\bar{v}_{t}^{(k)}$. Utilizing \eqref{eq:normalized_weight_variation}, %. As mentioned in previous section, assuming that $\|\nabla{\mathcal{L}_{acc}(w^k_{t})}\|$ does not vary so much \ro{during an epoch},
we can set $\bar{v}_{t}^{k} \approx \frac{\bar{\eta}^{k}}{n_k} \|\nabla{\mathcal{L}_{acc}(w^k_{t})}\|$, which
%, along with \eqref{eq:normalized_weight_variation}, 
leads to 
\begin{eqnarray}
\label{eq:new_learning_rate}
\bar{v}_{t}^{k}-v_{t}^{k} \approx \frac{\bar{\eta}_{t}^{k}-\eta_{t}^{k}}{n_k} \|\nabla{\mathcal{L}_{acc}(w^k_{t})}\|=\frac{\bar{\eta}_{t}^{k}-\eta_{t}^{k}}{\eta^k}v_{t}^{k}.
\end{eqnarray}
Finally we can get the renewed LR as
\begin{eqnarray}
\label{eq:autotuning_learning_rate}
%  \bar{\eta}_{t}^{k} \approx \frac{\eta_t^k (\bar{v}_{t}^{k}-v_{t}^{k})}{v_{t}^{k}} + \eta_{t}^{k}.
 \bar{\eta}_{t}^{k} \approx \frac{\eta_t^k \bar{v}_{t}^{k}}{v_{t}^{k}}.
\end{eqnarray}
In actual AutoLR scheme, convergence to the goals of each epoch cannot be done at once, so we adopt an iterative trial policy as follows:
\begin{eqnarray}
\label{eq:autotuning_learning_rate2}
 \bar{\eta}_{t}^{k} \gets \frac{\eta_t^{k(i)} \bar{v}_{t}^{k}}{v_{t}^{k(i)}} , ~~\eta_{t}^{k(0)}=\eta_{t}^{k},
\end{eqnarray}
where $i$ is the trial index.
This renew procedure is repeated until the goal \eqref{eq:ordering_of_variation} is achieved in each epoch. The overall flow of AutoLR is given in Algorithm
\ref{alg:adjustLR}.

\renewcommand{\algorithmicrequire}{\textbf{Auto-tuning:}}
\renewcommand{\algorithmicensure}{\textbf{Notation:}}
\renewcommand{\algorithmicprocedure}{\textbf{Auto-tuning:}}
\algtext*{EndWhile}
\algtext*{EndIf}
\algtext*{EndFor}

\algdef{SE}[DOWHILE]{Do}{doWhile}{\algorithmicdo}[1]{\algorithmicwhile\ #1}

\begin{algorithm} [t]
\caption{AutoLR: Auto-tuning of learning rates}\label{alg:adjustLR}
\small{
\begin{algorithmic}[1]
\Ensure 
\Statex $\eta^k$ : learning rate of $k$-th block
\Statex $\bar{\eta}^k$ : target learning rate of $k$-th block
\Statex $v_t^k$ : weight variation of $k$-th block in $t$-th epoch
\Statex $\bar{v}_t^k$ : target weight variation of $k$-th block in $t$-th epoch
\Statex \textit{network} : network with tuned weights
\Statex \textit{trial-network} : trial network for tuning of $\eta^k$
%\Statex Set \textit{goal} to (2) : set target weight variation for each block.
%\Statex 
\Require 
\State Initialize $\{\eta_k\}$ with $\{\eta_k^0\}$
\State Initialize \textit{network} with \textit{pre-trained network}
\State Set \textit{sorting quality} $=0$.        

%\Statex 
\For {epoch $\gets 1$ to $T$} 
    \State \textit{trial-network} $\gets$ \textit{network} 
    \While {\textit{sorting quality} $\leq \tau_{s}$}
        \State Fine-tune \textit{trial-network} for target dataset
        \State Calculate \textit{weight variation} in (1)
        \State Calculate \textit{sorting quality} in (9)  
        \If {\textit{sorting quality} $>\tau_{s}$}
             \State \textit{network} $\gets$ \textit{trial-network}
        \Else 
            \If {epoch $ == 1$} 
                \State Renew $\bar{v}_t^k$ by (11)
            \Else
                \State Renew $\bar{v}_t^k$ by (12) and (13)
            \EndIf
            \State Renew $\bar{\eta}^k$ by (15)
            \State ${\eta}^k \gets \bar{\eta}^k$
        \EndIf
    \EndWhile 
    \State epoch++
\EndFor
\end{algorithmic} }
\end{algorithm}

Note that, the guarantee of convergence for Eq.\eqref{eq:autotuning_learning_rate2} and the guidance of setting for $\alpha$ and $\beta$ are included in the supplementary material.
The hyper-parameters $\alpha$ and $\beta$ in our algorithm effectively saves the effort of searching through too much combinations of continuous real spaces to find the appropriate learning rate for each layer. In all experiments, $\alpha$ and $\beta$ were set to 2 and 0.4 empirically following the guidance in the supplementary material. 
The threshold parameter $\tau_{s}$ was loosely set to 0.94 so that the training speed was not too slow by an approximate sorting  instead of the complete sorting by setting $\tau_{s}=1$.

\section{Experiments}
This section shows our experimental results. 
We begin by describing the target datasets and metric. Then, the experimental details are presented.
We then describe the ablation study and visualization results of our algorithm. 
Finally, we show the
comparison results with existing methods
in the remaining sections.

\subsection{Datasets}
\label{dataset}
\textbf{CUB-200}~\cite{ref:CUB-2011} dataset consists of 200 different species of birds. The first 100 classes are used for training and the other 100 classes for testing.\\
\textbf{Cars-196}~\cite{ref:stcar}
 dataset has 196 categories of car images and its total number is 16,185. The first 98 classes are used for training an the other 98 classes for testing. \\
\textbf{Stanford Online Products (SOP)}~\cite{ref:SOP}
 has 120,053 images of 22,634 categories of products. 11,318 and 11,316 classes are used for training and testing, respectively.\\
\textbf{Inshop}~\cite{ref:D_inshop}
 has 54,642 images of 11,735 categories of clothing items. 3,997 classes are used for training and the other 3,985 classes for testing. 
In the case of Inshop, the test dataset is divided into query and gallery. 

\subsection{Evaluation metric}
The Recall@K metric~\cite{ref:SOP} is employed for evaluating compared methods in image retrieval task.

\subsection{Implementation Details}
\label{Implementation detail}
% All experiments were implemented and conducted pytorch with Geforce GTX 1080 Ti machine.
We utilized ResNet-50~\cite{ref:resnet} pre-trained by using ImageNet~\cite{ref:imagenet}.
The input size was set to 224$\times$224 and the batch size was set to 40.
% The hyper-parameters $\alpha$ and $\beta$ for setting a range of weight variation were empirically set to 0.4 and 2.
% The threshold parameter $\tau_{s}$ was set to 0.94. 
For input data augmentation, the horizontal flipping with 0.5 probability was employed in training.
The initial LR was set to 1e-3.
We used the cross-entropy loss function. 
The stochastic gradient descent (SGD) optimizer was used along with Nesterov momentum~\cite{ref:nesterov}. Initial momentum rate and weight decay coefficient were set to 0.9 and 5e-4, respectively. 
%Because the labels of testing images are totally different from training images, the classifier that using for training images is ejected for testing.
\begin{figure*}[t]
\centering
{\includegraphics[width=17cm]{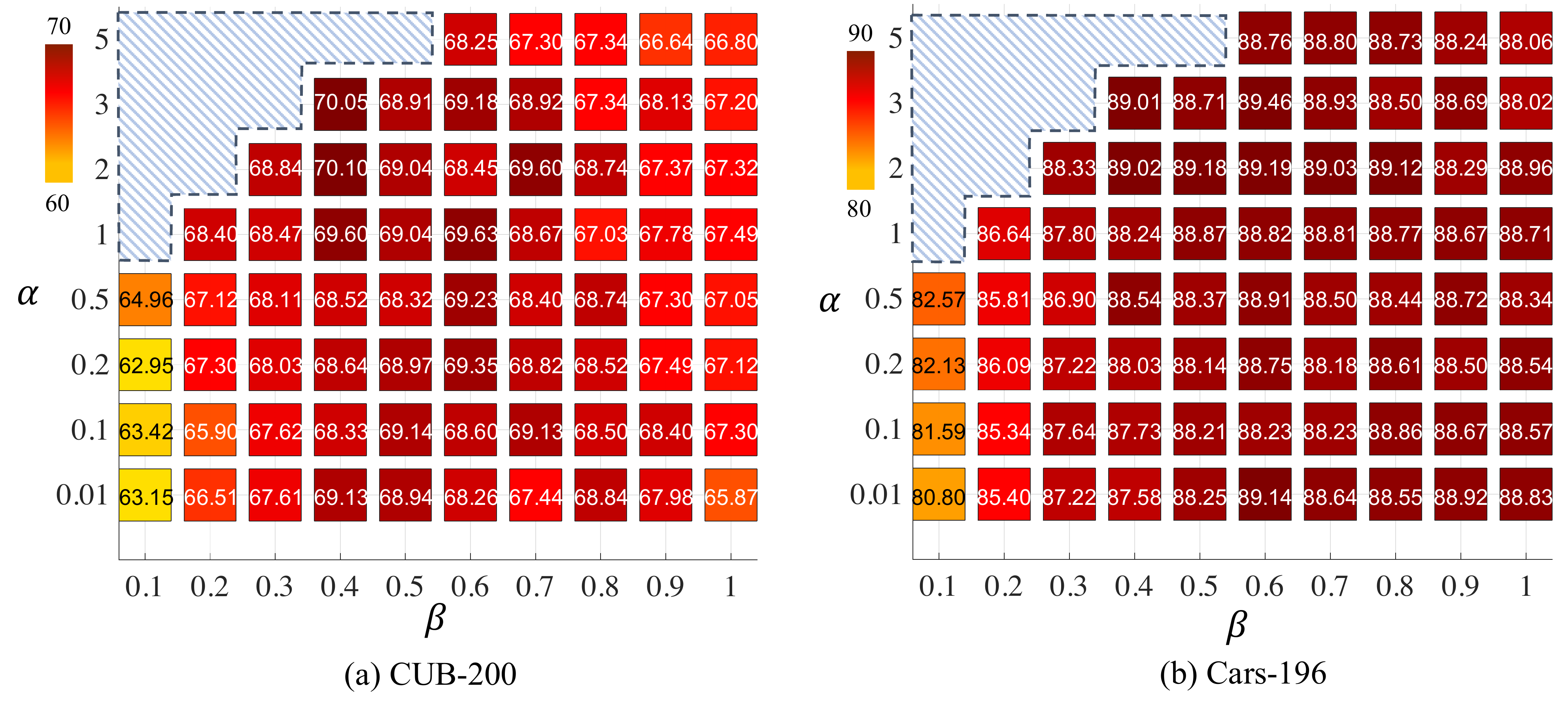}}
    % \hfill    
    \vspace{-0.6cm}
    \caption{Recall@1 results according to the hyper-parameters $\alpha$ and $\beta$ on CUB-200 and Cars-196}
    \label{fig:hyper_param}
    \vspace{-0.2cm}
\end{figure*}

\begin{table*}[t!]
\caption{Recall@\textit{K} score of proposed method on image retrieval dataset for the ablation study}
% \vspace{+0.2cm}
\label{table:ablation}
% \footnotesize{
\resizebox{1\textwidth}{!}{
\centering
\begin{tabular}{l||cccc|cccc|cccc|cccc}
% \hline
\toprule
&\multicolumn{4}{c|}{CUB-200} &\multicolumn{4}{c|}{Cars-196} &\multicolumn{4}{c|}{SOP} &\multicolumn{4}{c}{InShop} \\
Variants    & 1	& 2  & 4  & 8  & 1	& 2  & 4  & 8  	& 1	& 10  & $10^2$  & $10^3$  	& 1	& 10  & 20  & 30     \\ %\hline
\midrule
(1)~SingleLR    & 63.88	& 75.14 & 83.86  & 90.26  & 85.89	& 91.58 & 95.12  & 97.45  	   & 80.02 	& 91.21  & 96.20 & 98.64        & 87.64 & 97.12  & 98.04 & 98.40      \\
(2)~Pruning only	    & 67.00	& 78.49	& 86.85	&92.07  & 87.85	& 93.11	& 96.06	& 98.02          & 83.31 	& 92.90  & 96.68 & 98.62    & 88.62 & 97.16  & 97.95 & 98.40     \\
(3)~AutoLR	  & \textbf{70.10}    & \textbf{80.62}   & \textbf{88.08}  & \textbf{92.98}    & \textbf{89.02}    & \textbf{94.23}   & \textbf{96.72}  & \textbf{98.14}	& \textbf{84.24}    & \textbf{93.47}   & \textbf{97.15}  & \textbf{98.92} 	& \textbf{91.72}    & \textbf{98.04}  & \textbf{98.66} & \textbf{98.93}   \\
% Cut-out$\dagger$ + Auto	  &69.35 &79.88 & 87.36& 92.66      &\textbf{84.14} &\textbf{93.36} & \textbf{96.88}& \textbf{98.80} &- &- & -& -  \\ %\hline
\bottomrule
\end{tabular}
}
 \vspace{-0.2cm}

\end{table*}

% 69.36 80.47 88.10 92.71

% & \textbf{70.41}    & \textbf{80.89}   & \textbf{88.22}  & \textbf{93.05}ㄴ

\subsection{Discussion of Hyper-parameters $\alpha, \beta$}
In the above, we have illustrated that the tuning of layer-wise LRs yields much improved performance. To manually tune the best layer-wise LRs, there are too much combinations of layer-wise LRs in a multi-dimensional continuous real space even when the lower bound and upper bound on the range of LRs are given. 
But our AutoLR can easily find the layer-wise LRs with only using the lower and upper bounds that are denoted by
$\alpha \times \min_k{v_{0}^{(k)}}$ and ${\beta}
\times \max_k {v_{0}^{(k)}}$, respectively. 

In this section, we provide the investigation results for the hyper-parameters  $\alpha$ and $\beta$ on the CUB-200 and Cars-196 datasets.
As shown in Figure~\ref{fig:hyper_param}-(a), the best performance on CUB-200 dataset is achieved when the $\alpha$ and $\beta$ are set to values around 2 and 0.4, respectively.
In the case of Cars-196 dataset, the setting $\alpha$ and $\beta$ to 2 and 0.4 provides relatively high performance as shown in Figure~\ref{fig:hyper_param}-(b) but the setting $\alpha$ and $\beta$ to 3 and 0.6 provides the best performance. 
Note that the hatched area of the Figure~\ref{fig:hyper_param} is the unavailable area where the lower bound becomes larger than the upper bound.

%%%%%%%%%%%%%%%%%%%%%%%%%%%%%%%%%%%%%%%%%%%%%%%%%%%%%%%%%%%%%%%%
\subsection{Ablation Study}
\label{ablation}
We conducted ablation studies to validate the components of the proposed algorithm. 
We consider three ablation variants: (1) use conventional fine-tuning with the same LR over all layers (SingleLR), (2) apply layer-wise pruning only, and (3) apply layer-wise AutoLR with (2).
The ablation studies were conducted on CUB-200, Cars-196, SOP, and Inshop with the pre-trained ResNet-50.
Layer-wise pruning was done using our proposed Algorithm~\ref{alg:pruning}. Then, for all the target tasks, the layer-wise pruning 15,16 showed the best performance.
The second row of Table~\ref{table:ablation} shows a consistent performance improvement over all target datasets in Recall@1. 
In the case of Inshop dataset, the performance does not improve much, but the pruning of two layers contributes to a reduction in the network complexity while maintaining comparable performance. 
The results show that our simple layer-wise pruning is an effective way to both improve performance and reduce network complexity.

\begin{figure}[t!]
\centering
{\includegraphics[width=8cm]{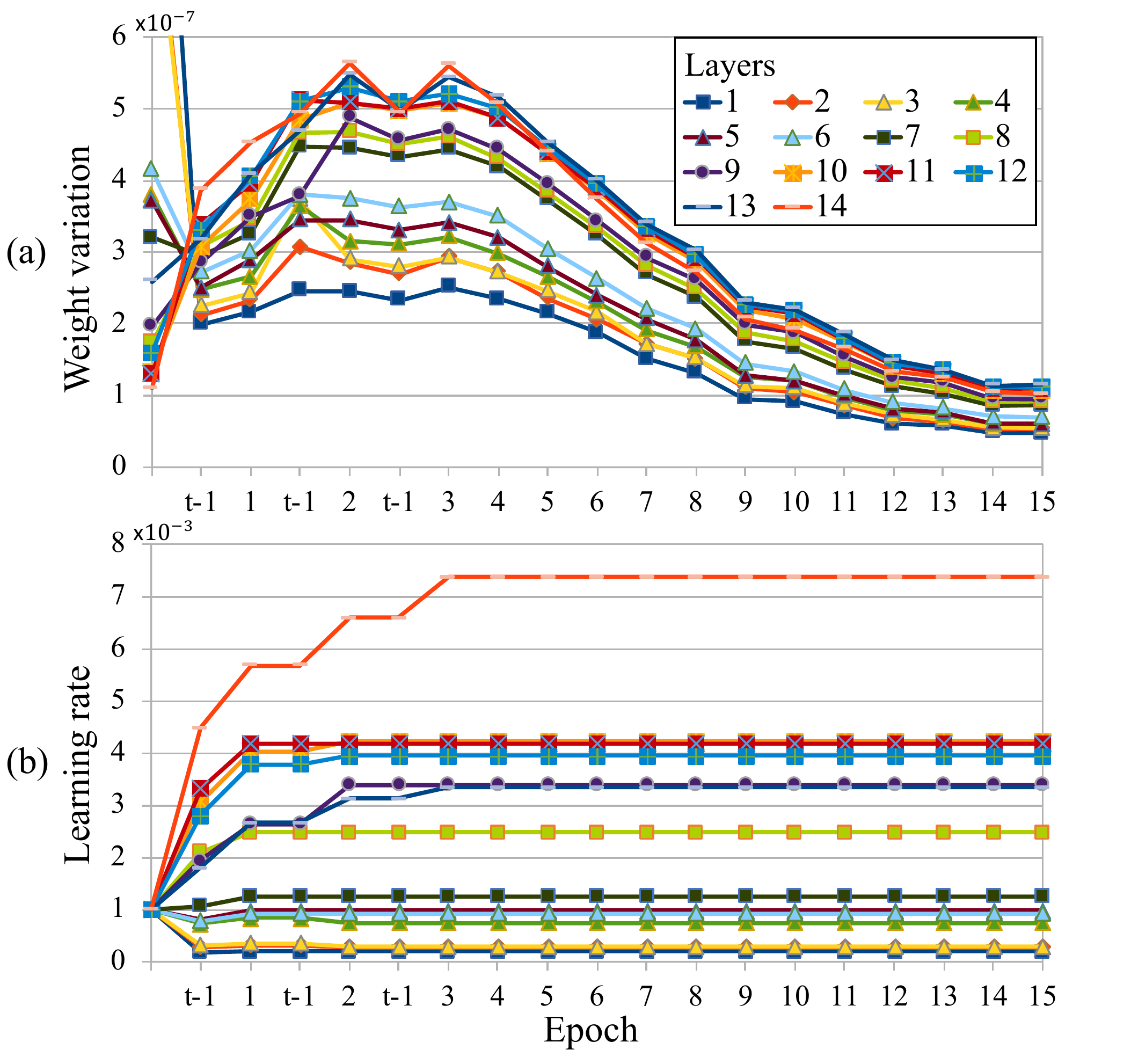}}
    \hfill    
    \vspace{-0.4cm}
    \caption{(a) Layer-wise weight variations and (b) layer-wise learning rate adaptations by our AutoLR algorithm}
    \label{fig:full_graph}
    \vspace{-0.4cm}
\end{figure}

The results for variant (3) are shown in the third row of Table~\ref{table:ablation}.
There are meaningful performance improvements for the three target tasks.
Figure~\ref{fig:full_graph} shows the layer-wise trends of weight variations and LRs by our algorithm for layer-wise AutoLR.
The learning was done up to 50 epochs, but after the convergence process was omitted after 15 epochs. $t$-$i$ represents the $i$-th automatic tuning trial in each learning epoch.
Figure~\ref{fig:full_graph}-(a) shows that our AutoLR algorithm achieves the goal that the magnitudes of the weight variations are sorted in ascending order from low-level to high-level layers. 
In Figure~\ref{fig:full_graph}-(a), Layer 14 is observed to be unsorted after four epochs. This is because the parameter $\tau_{s}$ to determine the successful sorting quality is not set to 1 (perfect sorting).
Figure~\ref{fig:full_graph}-(b) shows how the layer-wise LRs are adjusted from the initial LR of 0.001 for all layers and converge to layer-wise constants. 
The trial tuning iteration was performed once or twice before three epoch, and thereafter the first tuning was successful without further trial tuning. 
Hence the additional overhead for trial tuning iterations required by the proposed method is negligible.
According to the LR trends in the highest layer 14, its change in each tuning was the largest. This result supports our Hypothesis 2 that the high-level layer is specific to the target task and thus its weight variations should be large to adapt itself to the new target task. To meet the goal, the LR in Layer 14 converges to a large value promptly. Our AutoLR algorithm also is valid for other layers, as shown in Figure~\ref{fig:full_graph}-(b).

In conclusion, the ablation study illustrates that the proposed layer-wise pruning and AutoLR algorithm is an effective and promising ways to improve performance and reduce network complexity by setting constant $\alpha, \beta$ and $\tau_s$ regardless of datasets.

\begin{figure}[t!]
\centering
{\includegraphics[width=7.5cm]{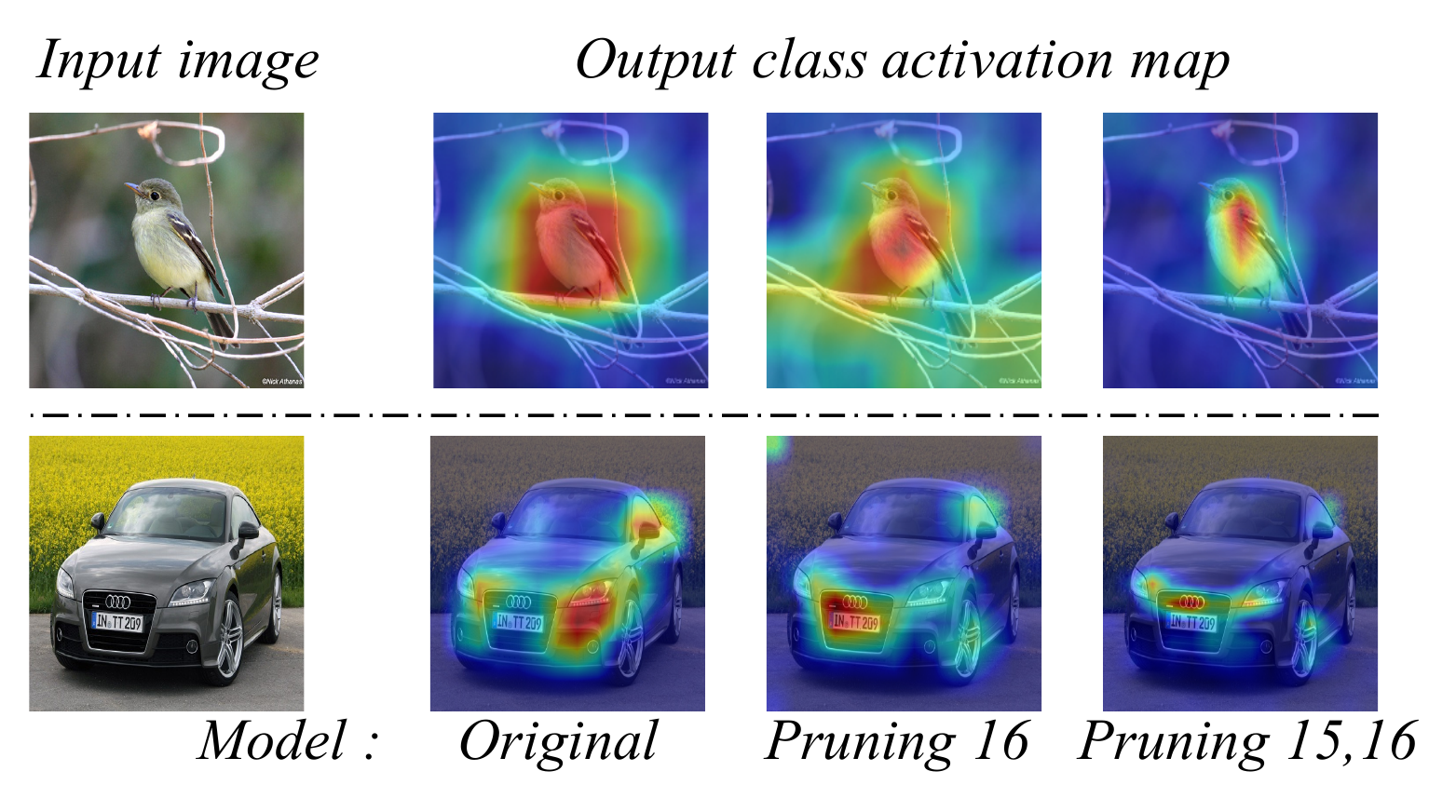}}
    \centering  
    \vspace{-0.3cm}
    \caption{The class activation map (CAM) visualization of the last layers by applying different layers pruning}
    \label{fig:grad_cam_prune}
\vspace{-0.2cm}
\end{figure}

% \begin{figure}[h]
% \centering
% {\includegraphics[width=7.5cm]{fig/cub_gradcam_until1415.pdf}}
%     \centering  
%     \vspace{-0.6cm}
%     \caption{The class activation map (CAM) visualization of the last layers by applying different layers pruning}
%     \label{fig:grad_cam_prune}
%     {\includegraphics[width=8.5cm]{fig/cub_gradcam2.pdf}}
%     \hfill    
%     \vspace{-0.6cm}
%     \caption{The class activation map (CAM) visualization of several layers (1, 4, 8, 14) according to the sorting quality. {\it initial-tune} is done by the conventional fine-tuning with single LR}
%     \label{fig:grad_cam_auto}
% \vspace{-0.4cm}
% \end{figure}
\begin{figure}[t!]
\centering
{\includegraphics[width=7.5cm]{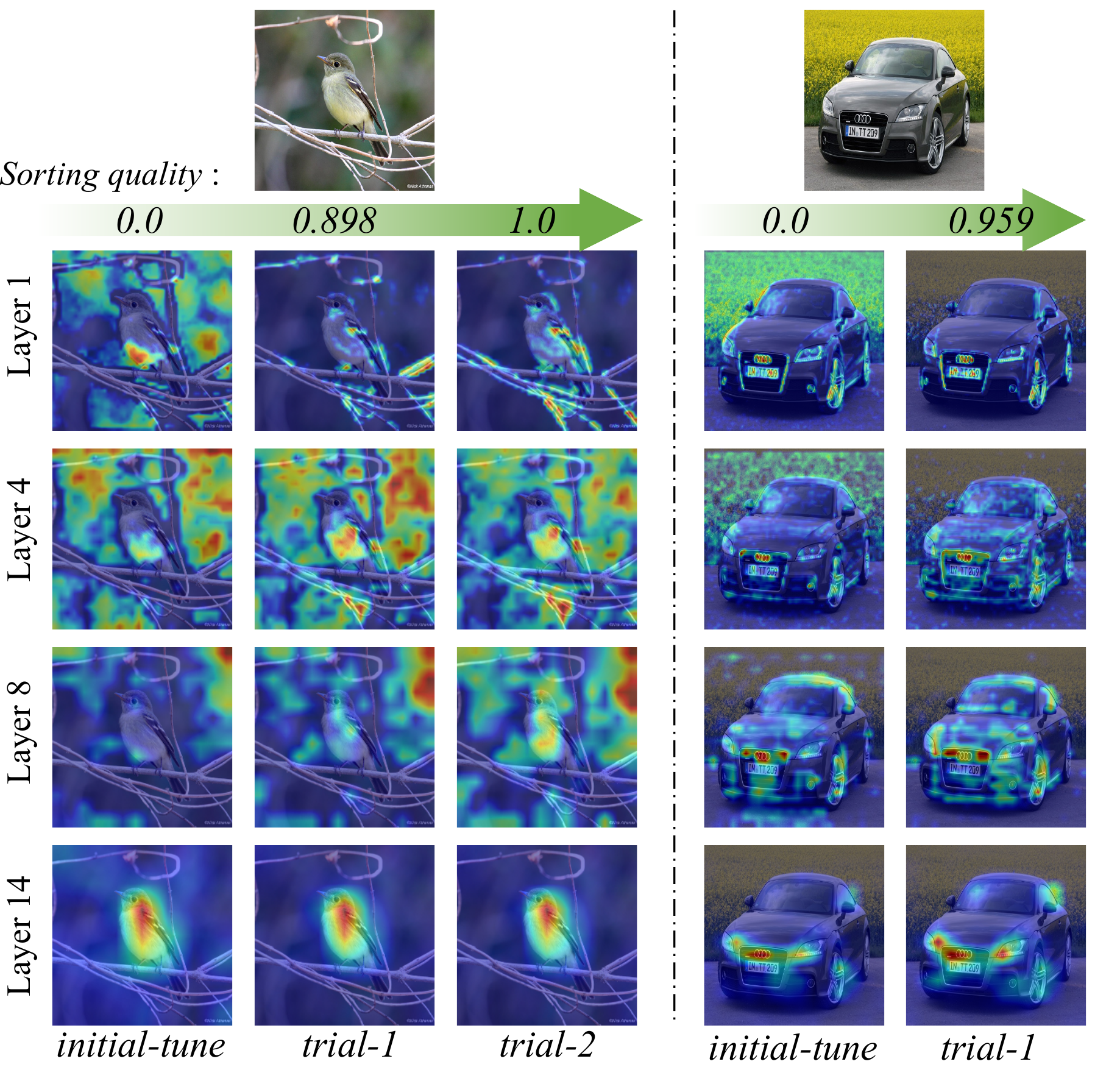}}
    \hfill    
    \caption{The class activation map (CAM) visualization of  layers (1, 4, 8, 14) according to the sorting quality. The {\it initial-tune} is done by the conventional fine-tuning with single LR}
    \label{fig:grad_cam_auto}
    \vspace{-0.4cm}
\end{figure}

%%%%%%%%%%%%%%%%%%%%%%%%%%%%%%%%%%%%%%%%%%%%%%%%%%%%%%%%%%%%%%

\subsection{Visualization on Effect to Layer-wise Features}
\label{visualization}

To understand how the sorting quality of layer-wise weight variations affects the responses in the layers, 
we investigated the class activation map (CAM) in each layer using a visualization technique, Grad-CAM~\cite{ref:grad_cam}. 

Figure~\ref{fig:grad_cam_prune} shows the CAM at the last layer of each pruned model.
In the case of \textit{Original} without pruning, activation gives attention to a relatively large area. However, as  high-level layers are pruned one by one, activation has more attention to the object or specific area, although the receptive field of each pruned one is the same. This supports our Hypothesis 1 that the there may be useless high-layers of a pre-trained network in a new task.

\begin{table}[t!]
\centering
\caption{Recall@K score of proposed Layer-wise pruning applied to Inception-V3 on CUB-200 dataset}
\label{table:inception_pruning}
\begin{tabular}{lccccc}
\toprule
Method     	& R@1	& R@2  & R@4  & R@8  \\ \midrule
Original & 54.12	&66.44	&76.54	&85.38  \\
Pruning 13   & 60.77	&71.94	&81.68	&88.93   \\
Pruning 12, 13   & 63.67	&74.71	&83.78	&90.60   \\
Pruning 11, 12, 13   & 60.72	&72.84	&82.85	&89.74 \\
\bottomrule
\end{tabular}
\end{table}

\begin{table*}[t]
\centering
\caption{Comparison with state-of-the-art methods on image retrieval datasets (Recall@\textit{K} Score)} %(ResNet-50 used, $\dagger$H means head of hydra structure)}
\label{table:sota_retrieval}
\vspace{+1mm}
% \footnotesize{
\resizebox{1\textwidth}{!}{
\begin{tabular}{ll||cccc|cccc|cccc|cccc}
\toprule
      & & \multicolumn{4}{c|}{CUB-200} & \multicolumn{4}{c|}{Cars-196} & \multicolumn{4}{c|}{SOP} & \multicolumn{4}{c}{Inshop} \\ 
Method   & Network  & 1     & 2  &4     &8  & 1     & 2  &4     &8   & 1     & 10 & $10^{2}$  &$10^{3}$ & 1     & 10  &20  &30\\ \midrule
%baseline& 89.3 &97.5  & 98.4          & 81.8	& 92.3  & 96.7      \\
A-BIER &Inception-v1 & 57.5 & 68.7 & 78.3 &86.2         & 82.0 & 89.0  & 93.2 & 96.1  & 74.2 & 86.9  & 94.0 & 97.8 & 83.1 & 95.1 & 96.9 &97.5  \\
DREML &Inception-v3 &58.9 & 69.6 & 78.4 &85.6     &-&-&-&-     &-&-&-&-      &78.4&93.7&95.8&96.7                                                    \\
ABE-8 &Inception-v1 & 60.6 &71.5  & 79.8 &87.4     & 85.2	& 90.5  & 93.9 &96.1    & 76.3	& 88.4  & 94.8 &98.2      & 87.3 &96.7  & 97.9& 98.2   \\             
NormSoft &ResNet-50  & 61.3& 73.9 & 83.5 & 90.0         & 84.2 & 90.4 & 94.4 & 96.9  & 79.5 & 91.5 & 96.7& -  & 89.4 & 97.8 & \textbf{98.7}& \textbf{99.0} \\
Margin &ResNet-50  &63.6 &74.4 &83.1  &90.0     & 79.6 & 86.5 & 91.9&  95.1      & 72.7 & 86.2 & 93.8&  98.0             &-&-&-&-                \\
MS &Inception-v1 & 65.7 & 77.0 & 86.4 & 91.2         & 78.2 & 90.5 & 96.0& \textbf{98.7}     & 78.2 & 90.5 & 96.0& 98.7       & 89.7 & 97.9 & 98.5& 98.8   \\ 
% SCHM &Inception-v1 & 66.2 & 76.3 & 84.1 & 90.1         & 83.6 & - & - & -     \\ 
DGCRL &ResNet-50 &67.9 &79.1 & 86.2 & 91.8 &75.9 &83.9 &89.7 &94.0  & - & -& -& -& -& -& -& -\\
Proxy-Anchor &ResNet-50 & 69.7 &80.0 & 87.0 & 92.4          & 87.7 & 92.9 & 95.8 & 97.9  & - & -& -& -& -& -& -& -\\ \midrule
% Proxy-Anchor &BN-Inception & 68.4 &79.2 & 86.8 & 91.6          & 86.1 & 91.7 & 95.0 & 97.3  & 79.1 &90.8 & 96.2 & 98.7          & 91.5 & 98.1 & \textbf{98.8} & \textbf{99.0}   \\ \midrule

\textbf{Pruning only}	  &ResNet-50   & 67.0	& 78.5	& 86.9	&92.1         & 87.9 	& 93.1  & 96.1 & 98.0    & 83.3 	& 92.9  & 96.7 & 98.6    & 88.6 & 97.2  & 98.0 & 98.4   \\
\textbf{AutoLR} &ResNet-50 & \textbf{70.1}    & \textbf{80.6}   & \textbf{88.1}  & \textbf{93.0}   	& \textbf{89.0}    & \textbf{94.0}   & \textbf{96.9}  & {98.4} 		& \textbf{84.2}    & \textbf{93.5}   & \textbf{97.2}  & \textbf{98.9}  	& \textbf{91.7}    & \textbf{98.0}  & \textbf{98.7} & 98.9    \\ \bottomrule
\end{tabular}}
 \vspace{-0.2cm}
\end{table*}

Figure~\ref{fig:grad_cam_auto} shows the CAM at each layer at the first epoch, where the sorting quality increases by AutoLR via one or two trial iterations. The {\it initial-tune} in Figure~\ref{fig:grad_cam_auto} is done using the conventional fine-tuning with single LR and the remaining trials are done using our AutoLR algorithm. 
As shown in Figure~\ref{fig:grad_cam_auto}, the CAM at each layer tends to have more attention to the object as the sorting quality increases by our AutoLR..
In Layer 1, as the sorting quality increases, unnecessary areas are deactivated and essential activation is formed in the target object area. 
This is because the AutoLR does not corrupt general features on the unnecessary area while the existing fine-tuning learns excessively the unnecessary area as a specific feature of the new target. 
In Layer 4, the CAM does not vary on unnecessary area; the CAM on the target area tends to be more attentive as the sorting quality increases. 
In Layer 8, the CAM on the target area tends to be more attentive as the sorting quality increases. However, the CAM also be attentive on unnecessary areas for the bird image due to the high LR tuned by our algorithm.  
In Layer 14, due to the pruning Layers 15 and 16 not being well-tuned by the existing fine-tuning, as shown in Figure \ref{fig:grad_cam_prune},  Layer 14 already has a good attention to the target. However, the activation is more attentive to the target as the sorting quality increases.

\subsection{Layer-wise pruning for other backbone}
To show the generality to other backbone networks, we conducted an experiment applying our layer-wise pruning scheme to Inception-V3. We divided Inception-V3 network into 13 layers as \{Conv2d(3), Conv2d(2), Mixed\_5b, Mixed\_5c, Mixed\_5d, Mixed\_6a, Mixed\_6b, Mixed\_6c, Mixed\_6d, Mixed\_6e, Mixed\_7a, Mixed\_7b, Mixed\_7c\}.
As shown in Table~\ref{table:inception_pruning}, pruning 12, 13 layers shows the best performance on the CUB-200 dataset.

\begin{table}[t]
\caption{Comparison of our AutoLR with the various learning rate scheduling method for the CUB-200 dataset with the equally pruned ResNet-50 (Recall@\textit{K} Score)}
\centering
\label{table:other_learning}
\resizebox{0.4\textwidth}{!}{
\begin{tabular}{@{}lcccc@{}}
\toprule
Method                        & hyper-parameters              & R@1 & R@2 & R@4 \\ \midrule
%\multirow{2}{*}{
Step-
& $t_{d}: 40, \gamma~$: 0.1 & {67.17} & 78.19 &86.19 \\
decay & $t_{d}: 40, \gamma~$: 0.2 & 67.35 & {78.14} &86.04 \\\midrule
\multirow{2}{0pt}{Cyclic} & \textit{cycle} : 5  & {67.54} & 78.70 &{86.73} \\
                              & \textit{cycle} : 7 & 68.55 &{79.17} &86.70 \\\midrule
\multirow{2}{0pt}{SGDR} & $n_{reset}$ : 8  & 68.18 & {79.34} &86.34  \\
                              &  $n_{reset}$ : 14 & {68.16} & 78.98 & {86.50}\\ \midrule
                            %   &  $n_{reset}$ :3    & 67.40 & 78.58 &86.77  \\\midrule
\textbf{AutoLR}             &  $\alpha:2, \beta:0.4 $    & \textbf{70.10} & \textbf{80.62} &\textbf{88.08}\\ \bottomrule
\end{tabular}}
\end{table}

\subsection{Comparison with Other LR Settings}
Our AutoLR algorithm belongs to the LR scheduling category.
Here, we show the superiority of our algorithm by a comparative study on the LR scheduling. 
The compared methods are `Step-decay'~\cite{ge2019step},
`Cyclic'~\cite{ref:cyclical}, and `SGDR'~\cite{ref:warmSGD}. 
Step-decay is the most widely used method in fine-tuning~\cite{ref:do_better_imagenet,ref:part_pooling_ECCV2018,ref:ro_rollback}.
It conducts LR decay by multiplying to all layers by a value $\gamma$ at a drop timing $t_{d}$. The initial LR of Step-decay is set to $l_{max}$.
Cyclic adjusts the LR between the max value $ l_{max} $ and min value $ l_{min} $ with periodic triangular waveform. The number of cycle is a hyper-parameter.
Similarly, SGDR adjusts the LR to decrease exponentially and LR is reset to its initial value $l_{max}$.
These resets are repeated multiple times with a hyper-parameter $n_{reset}$. 

For fair comparison, all methods were applied to the equally pruned network (pruning 15, 16), and all experiments were conducted equally for 50 epochs. 
The $l_{max}$ and $l_{min}$ for all experiments were set to 0.01 and 0.001, respectively.
Depending on the hyper-parameters of each method, we tested the performance in a range guided in each method and selected the best performance. All experimental results  are given in Table \ref{table:other_learning} of the supplementary material, where 
the tuned hyper-parameters are also listed for all the methods.
As shown in Table~\ref{table:other_learning}, our AutoLR algorithm outperforms the other LR scheduling methods consistently.

% \subsection{Validity on Fine-grained Classification}
% \label{fine-grained}
% To verify that our method is valid for other tasks, we conducted experiments on the fine-grained classification task.
% As shown in Table~\ref{table:fine-grained_sota}, our method outperforms Spot-tune~\cite{ref:spot-tune}, which is a similar method that  performs layer-wise tuning  with the same network ResNet-50.
% Spot-tune requires additional weights to determine layer-wise tuning, while our method reduces weights through layer-wise pruning. Nevertheless, our method performs better performance than Spot-tune. Hence, this result is quite meaningful.  In addition, as shown in Table~\ref{table:fine-grained_sota}, the proposed method using only fine-tuning achieves comparable performance to the state-of-the-art methods that use various kind of add-on techniques. 

%%%%%%%%%%%%%%%%%%%%%%%%%%%%%%%%%%%%%%%%%%%%%%%%%%%%%%%%%%%%%%

\subsection{Comparison with Existing Methods  }
\label{comp_sota}
Table \ref{table:sota_retrieval} shows that our AutoLR outperforms not only the methods of using the same backbone ResNet-50~\cite{ref:normsoft,ref:margin_retrieval,zheng2019towards,kim2020proxy}, but also the methods of using other backbone networks~\cite{ref:a-bier,ref:dreml,ref:attention_ensemble_ECCV2018,ref:Multi-Similarity}.
Another impressive one is that our method with only a pruning scheme outperforms the current SOTA in Cars-196 and SOP datsets even though it is a simple pruning method that removes just a couple of high-level layers assessed to be useless to a new task.

\section{Conclusion}
In this paper, we proposed a novel algorithm for layer-wise pruning and auto-tuning of layer-wise LRs with simple setting of hyper-parameters. The pruning algorithm uses a simple technique to prune a couple of high-level layers that are not helpful to a new task. 
The auto-tuning algorithm automatically adjusts the LRs depending on the role of each layer so that they contribute to performance improvement. 
The advantages of the proposed algorithm are not only simple for implementation, but also effective in improving performance and reducing network complexity.
The effectiveness and efficiency of the proposed algorithm has been validated by the experiments on four image retrieval benchmark datasets. 
Hence the proposed pruning and automatic tuning algorithm will be able to contribute to the advances for automated and efficient machine learning.

% 아래 내용 무난한 것 같은데, 이미 마감 끝났지?

% \section{Ethical Statement}
% We used only publicly accessible benchmark datasets for our algorithm development and evaluation. Therefore, there are no ethical issues regarding the data in our research. 
% In addition, each performance evaluation in this paper and comparison with other algorithms were conducted fairly and carefully, and supplementary materials were used to deal with hyper-parameter issues as much as possible.

% \input{input_tables/fine_grained_AutoLR}
% In this paper, we proposed a novel algorithm for layer-wise pruning and auto-tuning of layer-wise learning rates. The pruning algorithm can prune a couple of high-level layers which are not helpful to a new task in a simple way. The auto-tuning algorithm automatically adjusts the learning rates depending on role of each layer so that they contribute to improvement of performance.  
% To the best of our knowledge, the auto-tuning of layer-wise learning rates is the first attempt. In addition, the proposed layer-wise pruning is simple but efficient way to improve performance as well as to reduce network complexity. As validated on image retrieval and fine-grained classification tasks, the proposed algorithm shows promising features for automated and efficient machine learning. 

\section*{Acknowledgment}
This work was supported by the ICT R\&D programs of MSIP/IITP [No.B0101-15-0552, Development of Predictive Visual Intelligence Technology] and  [2017-0-00306, Outdoor Surveillance Robots].

\bibliographystyle{aaai}
\bibliography{egbib}

\end{document}